\documentclass[sigconf]{acmart}

\AtBeginDocument{%
  \providecommand\BibTeX{{%
    \normalfont B\kern-0.5em{\scshape i\kern-0.25em b}\kern-0.8em\TeX}}}

\setcopyright{acmcopyright}
\copyrightyear{2022}
\acmYear{2022}
\acmDOI{XXXXXXX.XXXXXXX}

\acmConference[ICONS '22]{International Conference on Neuromorphic Systems}{July 27--29,
  2022}{Oak Ridge, TN}

\usepackage[utf8]{inputenc}
\usepackage{graphicx} 
\usepackage{caption}
\usepackage{subfig}
\begin{document}

\title{Dictionary Learning with Accumulator Neurons}


\author{Gavin Parpart}
\email{gavin.parpart@pnnl.gov}
\author{Yijing Watkins}
\email{yijing.watkins@pnnl.gov}
\author{Carlos Gonz\'alez}
\email{carlos.gonzalezrivera@pnnl.gov}
\affiliation{%
  \institution{Pacific Northwest National Laboratory}
  \streetaddress{902 Battelle Blvd}
  \city{Richland}
  \state{WA}
  \country{USA}
  \postcode{99354}
}

\author{Edward Kim}
\email{ek826@drexel.edu}
\author{Jocelyn Rego}
\email{jr3548@drexel.edu}
\author{Andrew O'Brien}
\email{ao543@drexel.edu}
\author{Steven C. Nesbit}
\email{scn43@drexel.edu}
\affiliation{%
  \institution{Drexel University}
  \streetaddress{3141 Chestnut St}
  \city{Philadelphia}
  \state{PA}
  \country{USA}
  \postcode{19104}
}

\author{Terrence C. Stewart}
\email{terrence.stewart@nrc-cnrc.gc.ca}
\affiliation{%
  \institution{National Research Council of Canada}
  \streetaddress{University of Waterloo Collaboration Centre}
  \city{}
  \state{}
  \country{Canada}
  \postcode{}
}

\author{Garrett T. Kenyon}
\email{gkenyon@lanl.gov}
\affiliation{%
  \institution{Los Alamos National Laboratory}
  \streetaddress{P.O. Box 1663, MS-B256}
  \city{Los Alamos}
  \state{NM}
  \country{USA}
  \postcode{87545}
}

\renewcommand{\shortauthors}{Kenyon, Kim, Stewart, and Watkins, et al.}

\date{May 2022}

\begin{abstract}
The Locally Competitive Algorithm (LCA) uses local competition between non-spiking leaky integrator neurons to infer sparse representations, allowing for potentially real-time execution on massively parallel neuromorphic architectures such as Intel's Loihi processor. Here, we focus on the problem of inferring sparse representations from streaming video using dictionaries of spatiotemporal features optimized in an unsupervised manner for sparse reconstruction. Non-spiking LCA has previously been used to achieve unsupervised learning of spatiotemporal dictionaries composed of convolutional kernels from raw, unlabeled video. We demonstrate how unsupervised dictionary learning with spiking LCA (\hbox{S-LCA}) can be efficiently implemented using accumulator neurons, which combine a conventional leaky-integrate-and-fire (\hbox{LIF}) spike generator with an additional state variable that is used to minimize the difference between the integrated input and the spiking output. We demonstrate dictionary learning across a wide range of dynamical regimes, from graded to intermittent spiking, for inferring sparse representations of both static images drawn from the CIFAR database as well as video frames captured from a DVS camera. On a classification task that requires identification of the suite from a deck of cards being rapidly flipped through as viewed by a DVS camera, we find essentially no degradation in performance as the LCA model used to infer sparse spatiotemporal representations migrates from graded to spiking. We conclude that accumulator neurons are likely to provide a powerful enabling component of future neuromorphic hardware for implementing online unsupervised learning of spatiotemporal dictionaries optimized for sparse reconstruction of streaming video from event based DVS cameras.

\end{abstract}



\maketitle

\section{Introduction}

Spiking neural networks (SNNs) are computational models that mimic discrete binary output of biological neural networks. 
Compared with artificial neural networks (ANN), 
SNNs incorporate leaky-integrate-and-fire (LIF) dynamics that increases both algorithmic and computational complexity. Whereas the individual neurons in ANNs are often implemented as simple ReLU transfer functions that require no adjustment of  free parameters, SNNs often utilize LIF neurons whose time constants and firing thresholds must be carefully adjusted, traditionally by hand, to achieve the desired input/output behavior. 

The justification for such increased complexity is two fold. First, by using dedicated, potentially analog, circuit elements to instantiate individual neurons and by exploiting the low-bandwidth of event-based communication enabled by SNNs, such networks can be implemented in extremely low-power neuromorphic hardware \cite{Boahen_2017}, enabling real-time remote applications that depend on scavenged power sources such as solar recharge.
Second, there is evidence that biological neural circuits utilize spike timing to transmit information more rapidly and to dynamically bind distributed features via synchronous oscillations \cite{rullen_2001} \cite{thorpe_2001} \cite{stephens2006} \cite{kenyon2010}. 
The potential for mimicking the dynamics of biological neural networks in fast, low-power neuromorphic processors has motivated several efforts to develop such
devices \cite{Loihi2} \cite{davies2018loihi} \cite{Truenorth} \cite{SpiNNaker}.

Sparse coding accounts for a variety of experimentally measured linear and non-linear response properties of V1 simple cells \cite{ZhuMengchen2013} \cite{rozell2008sparse} \cite{paiton2020selectivity} \cite{teti2020can} and can be implemented in a biologically-plausible manner in terms of local competition using the Locally Competitive Algorithm \cite{rozell2008sparse}.
Algorithmically, sparse coding employs an overcomplete set of non-orthogonal basis functions (feature vectors) to infer a sparse combination of non-zero activation coefficients that most accurately reconstruct each input image. Most relevant for the present study, sparse coding provides a powerful and robust technique for the unsupervised learning of both static and spatiotemporal dictionaries for image and video classification, respectively.  Importantly, sparse coding  can be readily extended to the processing of streaming event data from DVS cameras by neuromorphic processors \cite{lundquist2016sparse} \cite{lundquist2017sparse} \cite{Watkins_icon2019} \cite{Watkins:2018:ICON} \cite{Kim_2017}.  
Given that DVS camera data is typically noisy and sparse coding is a powerful denoising tool  \cite{Watkins_icon2019} \cite{Watkins:2018:ICON}, we hypothesize that sparse coding may be ideal for inferring discriminative and robust representations from noisy event training.

This paper demonstrates an efficient procedure based on accumulator neurons \cite{accumulator_neurons} to greatly facilitate the interpolation between non-spiking and spiking neural networks (i.e. ANNs to SNNs). Accumulator neurons combine a conventional leaky-integrate-and-fire (\hbox{LIF}) spike generator with an additional state variable that is used to minimize the difference between the integrated input and the spiking output. Unlike conventional LIF models, accumulator neurons can produce multiple spikes on a single time step. Qualitatively, the behavior of an accumulator neuron is determined by an adjustable spike height. When the spike height is small in comparison to the change in the membrane potential produced by the synaptic input on a given time step, an accumulator neuron will fire multiple spikes so as to achieve maximum fidelity between input and output on that time step. When the spike height is large compared to the change in the membrane potential on a given time step, the difference between input and output "accumulates" until a single spike can approximately make up the difference. Accumulator neurons, thus, ensure that the input/output relationship remains approximately satisfied over time and that the overall dynamics of the neural system remains approximately unchanged even as the output of the individual neurons shift from graded to intermittent spiking.

Starting with LCA models employing non-spiking neurons, whose implementation has been extensively studied, we swap in accumulator neurons and increase the spike height so as to achieve progressively more realistic spiking behavior. Remarkably, we find that the resulting models remain able to learn spatial and spatiotemporal dictionaries, respectively, that support high-quality reconstructions as well as classification performance using a linear classifier that is competitive with what is achieved by a fully supervised deep CNN classifier.
Our results suggest that accumulator neurons can be an important enabling strategy for the online acquisition of adaptive behaviors by intelligent autonomous systems equipped with event-driven DVS cameras that provide sensory input to ultra-lightweight and low-power neuromorphic processors.


\section{Datasets}
\subsection{CIFAR-10}
The CIFAR-10 \cite{cifar10} is a standard dataset of labeled, thumbnail sized images used for benchmarking computer vision algorithms in the field of machine learning. The dataset is comprised of 60,000 32x32 RGB images from 10 classes. In this work, we center cropped the 16x16 images from the original images to reduce processing time.

\subsection{Poker-DVS}
\begin{figure}
\begin{center}
\includegraphics[width=0.44\textwidth]{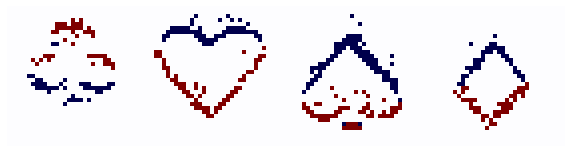}
\caption{Example frames for each Poker-DVS Class. Supra-threshold changes in local pixel intensity, corresponding to positive (red) and negative (blue) events, were summed over an 1 ms time window to produce a single frame.}
\label{fig:pokerdvs_input}
\end{center}
\end{figure}

The silicon retina of Dynamic Vision Sensor (DVS) camera is a form of imaging technology inspired by biological vision \cite{Lichtsteiner2008}.  A DVS camera only measures and transmits signed event data when the absolute value of a pixel's intensity changes beyond a predefined threshold. The resulting video resembles images run through an edge detection algorithm because light intensity changes tend to mostly occur at the edges (figure \ref{fig:pokerdvs_input}). This sensitivity to object boundaries allows the silicon retina camera to capture very fast dynamic events with relatively small bandwidth. 

The Poker-DVS dataset was obtained by fast browsing of a specially made poker card decks in front of a DVS camera for approximately two seconds \cite{serrano2015poker}. It includes extremely high-speed and noisy events across 35x35 pixels with 4 classes: clubs, diamonds, hearts and spades. 
For 48 training samples and 20 validation samples in this database, each card appeared on a DVS camera for about 20–30 ms. In this work, we accumulated the events for each card into 1 ms frames. Example frames for each class illustrated are in figure \ref{fig:pokerdvs_input}.




\section{Approach}
\subsection{Sparse Coding}
Given an overcomplete basis, sparse coding algorithms seek to identify the minimal set of generators that most accurately reconstruct each input image. In neural terms, each neuron is a generator that adds its associated feature vector to the reconstructed image with an amplitude equal to its activation. For any particular input image, the optimal sparse representation is given by the vector of neural activations that minimizes both image reconstruction error and the number of neurons with non-zero activity.  Formally, finding a sparse representation involves finding a minimum of the following cost function: 
\begin{equation}
E(\ \overrightarrow{I},\phi,\ \overrightarrow{a})=\min\limits_{ \{\overrightarrow{a}, \, \phi \} } \left[  \, \frac{1}{2}  ||  \overrightarrow{I} - \phi * \overrightarrow{a} ||^2 +	\lambda || \overrightarrow{a} ||_1\right],
\label{eq:SC}
\end{equation}
In Eq.~(\ref{eq:SC}), $\overrightarrow{I}$ is an image unrolled into a vector, and $\phi$ is a dictionary of feature kernels that are convolved with the sparse representation $\overrightarrow{a}$.  The factor $\lambda$ is a tradeoff parameter; larger $\lambda$ values encourage greater sparsity (fewer non-zero coefficients) at the cost of greater reconstruction error. Both the sparse representation $\overrightarrow{a}$ and the dictionary of feature kernels $\phi$ can be determined by a variety of standard optimization methods \cite{rozell2008sparse}. 

Our approach to compute a sparse representation for a given input image is based on a convolutional generalization of a rectifying Locally Competitive Algorithm (LCA)  \cite{rozell2008sparse}. Once a sparse representation for a given input image has been found, the basis elements associated with non-zero activation coefficients are adapted according to a local Hebbian learning rule that further reduces the remaining reconstruction error. 


LCA finds a local minimum of the cost function defined in Eq.~(\ref{eq:SC}) by introducing the dynamical variables (membrane potentials) $\mathbf{u}$ such that the output~$\mathbf{a}$ of each neuron is given by a soft-threshold transfer function, with threshold~$\lambda$, of the membrane potential\cite{Olshausen08}:

\begin{equation}
 \mathbf{a} = \begin{cases} 
      u -  \lambda,  & \mathbf{u} > \lambda \\
      0,  & otherwise
   \end{cases}
\end{equation}

The cost function defined in equation (\ref{eq:SC}) is then minimized by taking the gradient of the cost function with respect to $\mathbf{a}$ and solving the resulting set of coupled differential equations for the membrane potentials $\mathbf{u}$: 

\begin{equation}
\mathbf{\dot{u}} \propto -\dfrac{\partial E}{\partial \mathbf{a}} = -\mathbf{u} + \mathbf{\Phi}^T \{\mathbf{I} - \mathbf{\Phi}T_\lambda(\mathbf{u})\} +  T_\lambda(\mathbf{u}).
\label{eq:SC1}
\end{equation}

An update rule for feature kernels can be obtained by taking the gradient of the cost function with respect to $\mathbf{\Phi}$:

\begin{equation}
{\Delta}\mathbf{\Phi} \propto -\dfrac{\partial E}{\partial \mathbf{\Phi}} = \mathbf{a} \otimes \{I - \mathbf{\Phi} \mathbf{a}\} = \mathbf{a} \otimes \mathbf{R}
\label{eq:phi}
\end{equation}
where we introduced an intermediate residual layer $\mathbf{R}$ corresponding to the sparse reconstruction error.

\begin{figure}
\begin{center}
\includegraphics[width=4.275cm,height=4.8cm]{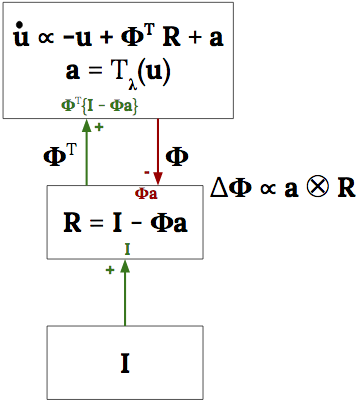}
\captionof{figure}{\footnotesize 
A LCA model that supports unsupervised dictionary learning via a residual or sparse reconstruction error layer.}
\label{modified_LCA}
\end{center}
\end{figure}

For non-spiking LCA, online unsupervised dictionary learning is achieved via a two step process:
First, a sparse representation for a given input is obtained by integrating Eq.~(\ref{eq:SC1}), after which Eq.~(\ref{eq:phi}) is evaluated to slightly reduce the reconstruction error given the sparse representation of the current input.

As illustrated in Figure~\ref{modified_LCA}, the weight update~(\ref{eq:phi}) resembles a local Hebbian learning rule for~$\mathbf{\Phi}$ with pre- and post-synaptic activities $\mathbf{a}$ and $\mathbf{R}$ respectively. However, the computation of  $\mathbf{\Phi}^T$ as well as the normalization constraint renders the overall dictionary learning process a non-local operation.

\subsection{Accumulator Neurons}

Accumulator neurons allow for a gradual transition from non-spiking to spiking regimes \cite{accumulator_neurons}. Accumulator neurons were originally developed so that one could train a standard non-spiking ANN using backprop and then gradually adjust the network towards a spiking domain using the trained weights.  Importantly, throughout the transition from non-spiking to spiking output, the average output of the accumulator neurons stays constant.

The core idea is to think of spiking as a discretization process, which then accumulates the error due to each discretization (i.e. the rounding error) onto the next time step. For example, consider a non-spiking neuron whose output varies between 0 and 1 (such as a sigmoid neuron).  If we discretize the output to steps of size 1, then the only outputs will be 0 or 1.  If the sigmoid neuron would output 0.5, then the discretized version could output a 0 on the first timestep, and a 1 on the next timestep, and then continue alternating.  In this way the average output continues to be 0.5, even though it is now a spiking neuron.

The way constancy of output is achieved is to take the error  in the output (i.e. the difference between the original neuron's output and the actual discretized output) and add it to the target output on the next timestep.  Importantly, we can adjust the size of the discretization steps, using a parameter $s$ ($1/\omega$ in \cite{accumulator_neurons}) denoting the size of each step in the output (i.e. if the allowed outputs are 0, 0.1, 0.2, 0.3, etc. then $s=0.1$).  If we interpret each step as a spike, then we can think of this as a spiking neuron that may fire more than one time per timestep, and each firing denotes an output of $s$).  For this reason, we can think of $s$ as a  ``spike height''.

\begin{figure}
\begin{center}
\includegraphics[width=8cm,height=6cm]{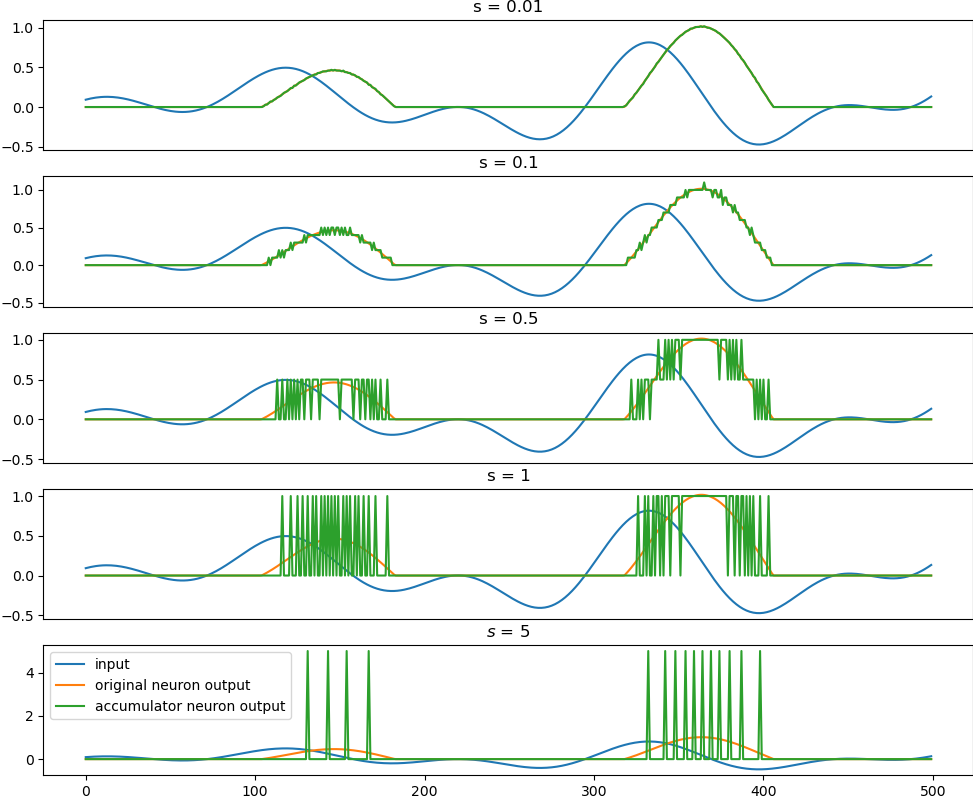}
\captionof{figure}{\footnotesize 
Top to bottom. spike height discretization factor ($s$) = 0.01, 0.1, 0.5, 1, 5. The example shows a gradual transition from non-spiking to intermittent spiking output using accumulator neurons with corresponding spike heights given by $s$.  }
\label{spike_height}
\end{center}
\end{figure}

This simple mechanism has a few features worth pointing out.  First, as $s$ approaches 0, the behaviour approaches that of whatever original non-spiking neuron model we started from.  Second, if $s$ is the maximum output value from the original neuron (and if the neuron always outputs a non-negative number), then we will have a neuron that will only spike at most once per timestep (i.e. a standard spiking neuron).  For values in-between, we can either think of the neuron spiking multiple times per timestep, or we can think of it as a single spike but the magnitude of that spike takes on a few different discrete values.  Third, if $s$ is much larger than the maximum output, then the neuron will still only spike at most once per timestep, but will also have much more temporal sparsity (for example, if $s=10$ for a sigmoid neuron, the neuron will spike at most once every 10 timesteps).  The $s$ parameter thus allows for trading off sparsity and accuracy, while still always maintaining the same average output from the neuron.   Examples showing this gradual tradeoff as spike height discretization factor $s$ is adjusted are shown in Figure~\ref{spike_height}.

Some neuromorphic hardware platforms such as SpiNNaker \cite{SpiNNaker} and Loihi 2 \cite{Loihi2} offer direct support for discretized spike magnitudes, so elements functionally similar to accumulator neurons may be efficiently supported even with intermediate $s$ values where multiple spikes occur per timestep.

\section{Results}

We evaluated unsupervised dictionary learning for sparse reconstruction and classification performance using both non-spiking and spiking LCA models, employing rate-coded and accumulator neurons, respectively, on both CIFAR10 static images and video frame captured from Poker-DVS events.
We implemented both LCA and S-LCA in Nengo \cite{nengo}, an open source neural simulation toolbox for building large-scale functional brain models, as well as in PyTorch \cite{Pytorch}.
By gradually transitioning from a non-spiking to a spiking LCA model, we demonstrate that dictionary learning supports both sparse reconstruction and classification performance that remains approximately constant despite the enormous range of dynamical regimes considered.  For large spike heights, corresponding to more spike-based output, we applied a low pass filter to the sparse latent representations to temporally average over individual spike events.

%
\vfill\null

\subsection{CIFAR-10}

\subsubsection{Unsupervised Dictionary Learning}
\hfill \\

\begin{figure}[htp]
\centering
\subfloat[Initial state.\label{fig:1a}]{\includegraphics[width=0.15\textwidth]{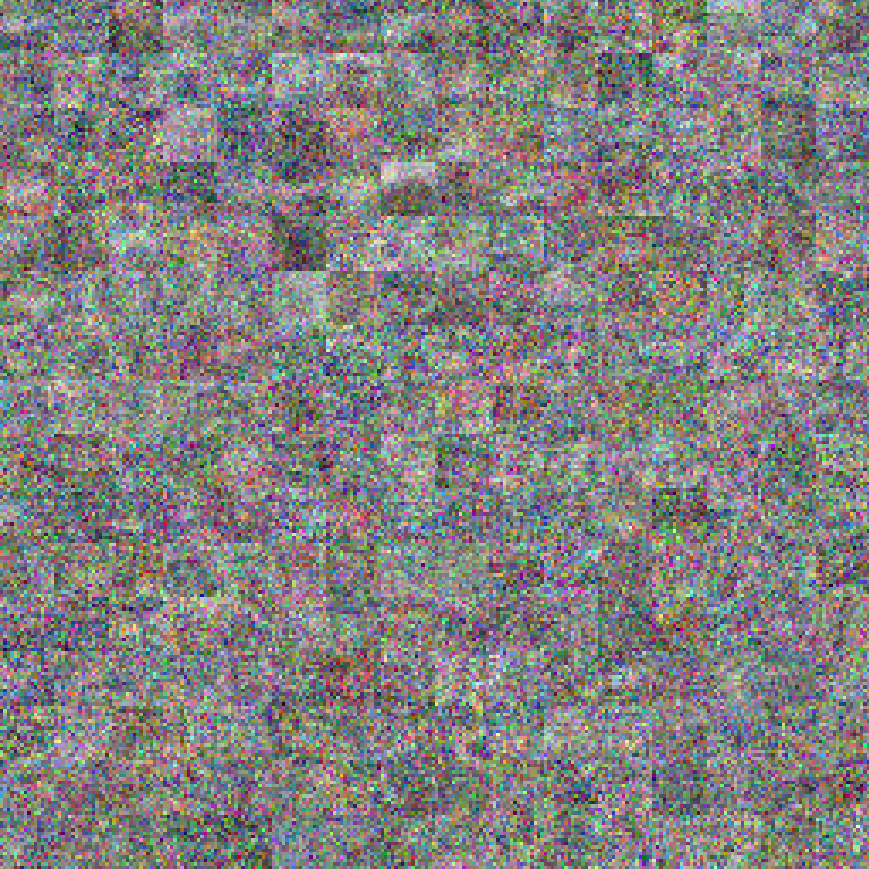}}\hfill
\subfloat[Middle state.\label{fig:1b}] {\includegraphics[width=0.15\textwidth]{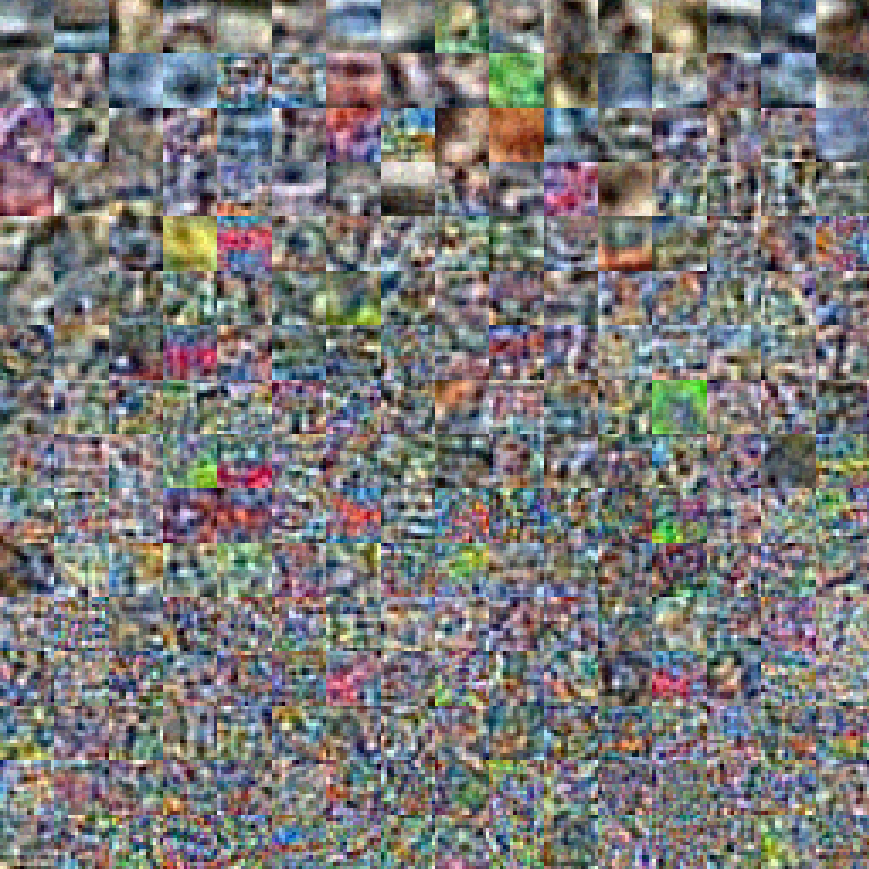}}\hfill
\subfloat[Final state.\label{fig:1c}]{\includegraphics[width=0.15\textwidth]{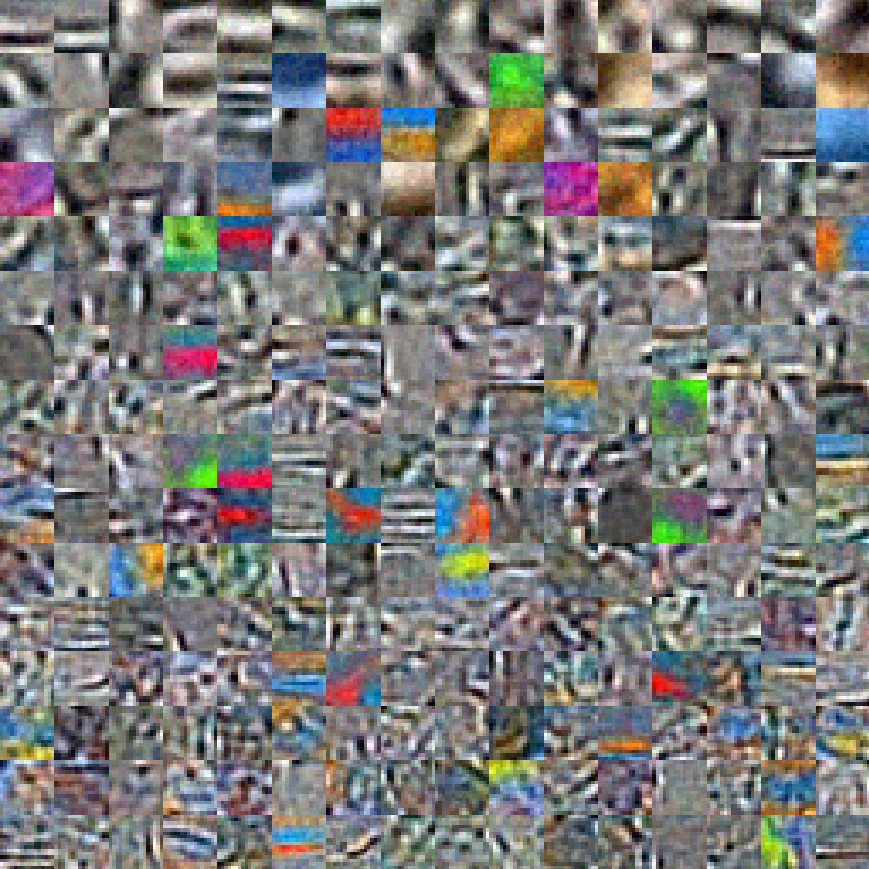}}
\caption{Top 256 most active dictionary elements using rate-coded neurons.} \label{fig:cifar_original_lca_dic}
\end{figure}

We demonstrate how unsupervised dictionary learning using spiking LCA (\hbox{S-LCA}) can be accomplished by starting with rate-coded non-spiking neurons and employing accumulator neurons to transition the model to a spiking regime. LCA allows for both unsupervised dictionary learning and inference to be performed in a manner compatible with the constraints of recent architectures, such as  the Intel Loihi research chip \cite{Loihi2}.

Non-spiking \hbox{LCA} was implemented using leaky integrator neurons with a ReLU-like transfer function with a finite threshold. \hbox{S-LCA} models were implemented using  accumulator neurons with spike height of 4.
Both \hbox{LCA} and \hbox{S-LCA} models consisted of a single inference layer containing 3840 features($\simeq5$ times overcomplete).
We initialized both models' dictionaries by sampling from a normal distribution with mean 0 and variance 0.01, and then rescaled each dictionary element to have unit L2 norm. We used center cropped $16\times16$ pixels CIFAR-10 images as input, where the input spike firing rate for each pixel is proportional to the pixel value. 

\begin{figure}[htp]
\centering
\subfloat[Initial state.\label{fig:cifar_spike_1a}]{\includegraphics[width=0.15\textwidth]{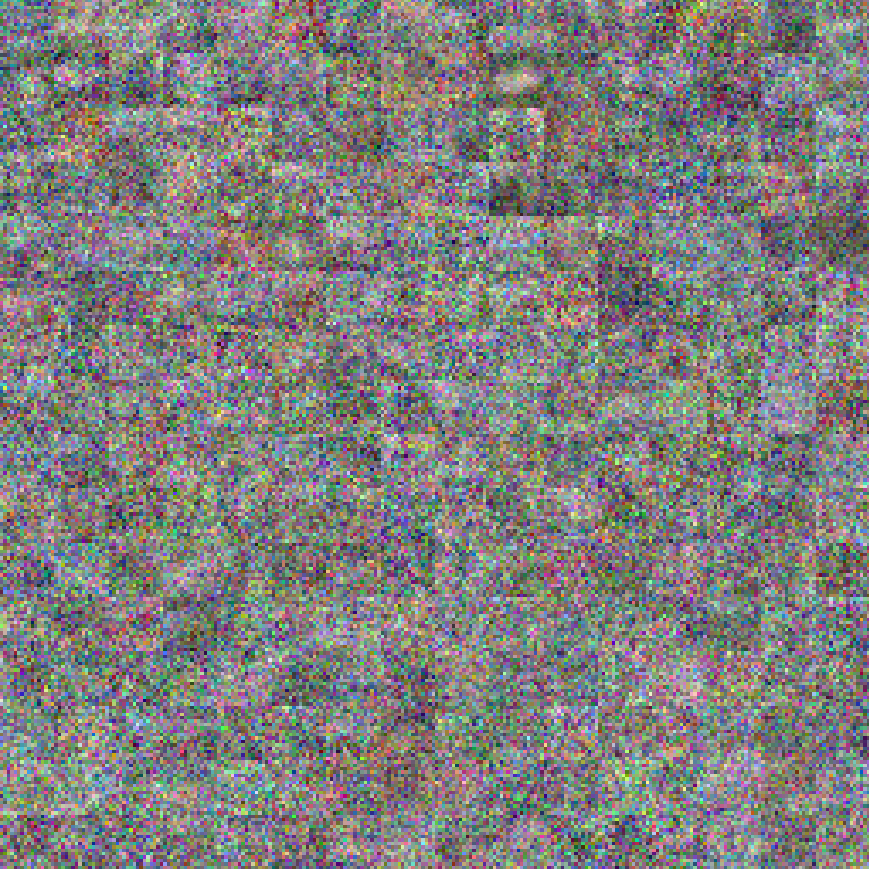}}\hfill
\subfloat[Middle state.\label{fig:cifar_spike_1b}] {\includegraphics[width=0.15\textwidth]{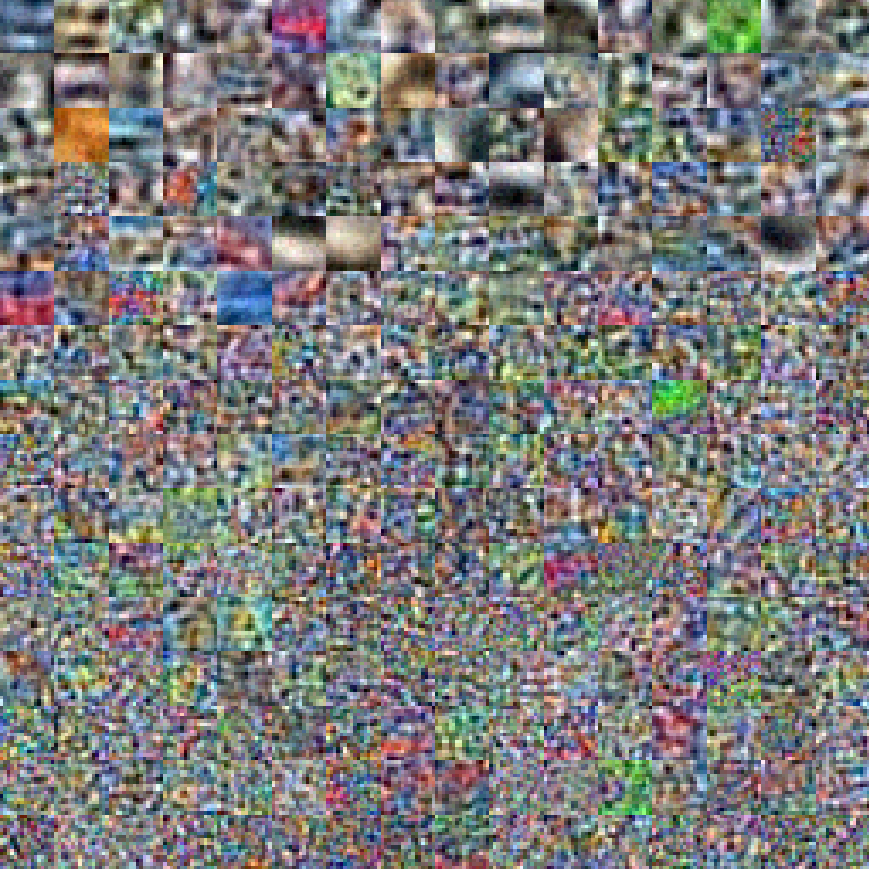}}\hfill
\subfloat[Final state.\label{fig:cifar_spike_1c}]{\includegraphics[width=0.15\textwidth]{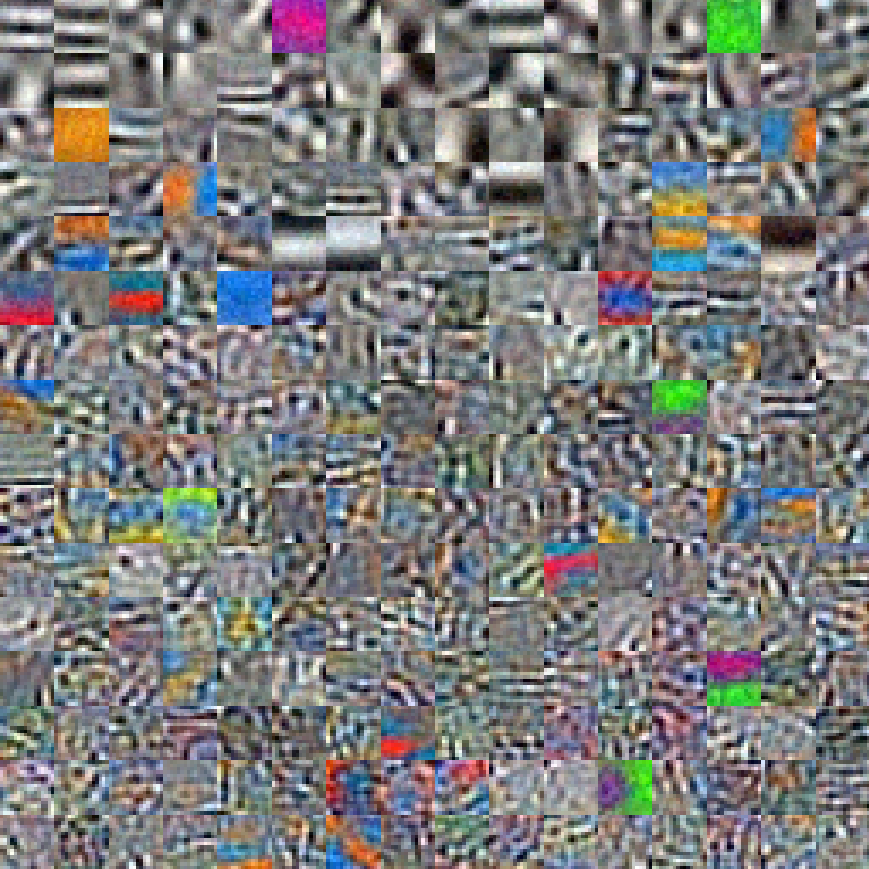}}
\caption{Top 256 most active dictionary elements using accumulator neurons with $s$ = 4.} \label{fig:cifar_spike_height_4_dic}
\end{figure}

The evolution of the dictionary learning using rate-coded neurons over one epoch is illustrated in Figure \ref{fig:cifar_original_lca_dic}.
The evolution of the dictionary learning using accumulator neurons with spike height set to 4 over one epoch is illustrated in Figure \ref{fig:cifar_spike_height_4_dic}.
Despite the enormous differences in dynamics between the two models, dictionary learning was largely unaffected.

\begin{figure}[htp]
\centering
\subfloat[Low activation .\label{fig:cifar_spike_1a}]{\includegraphics[width=0.15\textwidth]{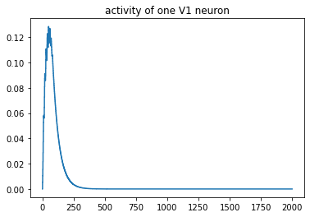}}\hfill
\subfloat[Median activation.\label{fig:cifar_spike_1b}] {\includegraphics[width=0.15\textwidth]{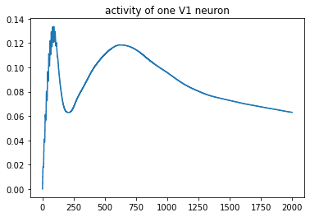}}\hfill
\subfloat[High activation.\label{fig:cifar_spike_1c}]{\includegraphics[width=0.15\textwidth]{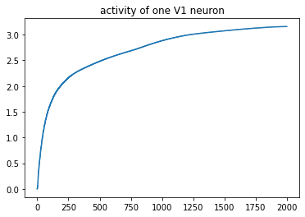}}
\caption{Activation of representative neurons in the LCA model over one display period.} \label{fig:cifar_spike_height_4_neurons}
\end{figure}

\begin{figure}[htp]
\centering
\subfloat[Low activation .\label{fig:cifar_spike_1a}]{\includegraphics[width=0.15\textwidth]{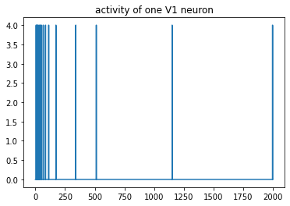}}\hfill
\subfloat[Median activation.\label{fig:cifar_spike_1b}] {\includegraphics[width=0.15\textwidth]{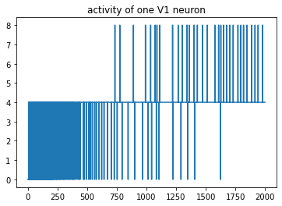}}\hfill
\subfloat[High activation.\label{fig:cifar_spike_1c}]{\includegraphics[width=0.15\textwidth]{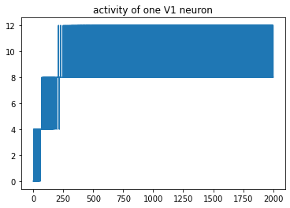}}
\caption{The activation of representative neurons in the S-LCA model with $s$ = 4 over one display period.} \label{fig:cifar_original_neurons}
\end{figure}

Figure~\ref{fig:cifar_original_neurons} shows representative neurons with different activation levels in the \hbox{LCA} model over a single display period (2000 time steps or 2s). The least active representative neuron was only above threshold at the beginning of the display period. On the other hand, the most active representative neurons were present with gradually increased activation over the display period.  Note the different activity scales on the y-axis.
Figure~\ref{fig:cifar_spike_height_4_neurons} shows representative activations in the \hbox{S-LCA} models using accumulator neurons with spike height = 4 over one display period. At spike height of 4, the least active accumulator neuron fires less than one spike per time step on average. In contrast, the most active accumulator neuron fires up to three spikes per time step. 

\subsubsection{Sparse Reconstructions}
\hfill \\

\begin{figure}[htp]
\centering
\subfloat[Inputs.\label{fig:cifar10_recons_1a}]{\includegraphics[width=0.45\textwidth]{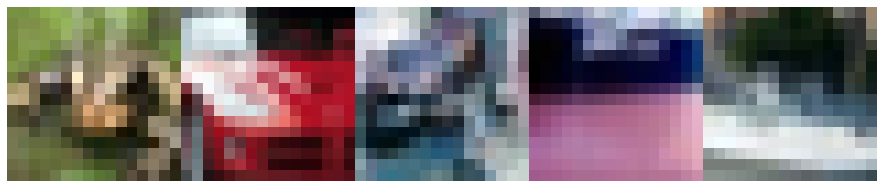}}\hfill
\subfloat[LCA Reconstructions. RMSE = 0.0195.\label{fig:cifar10_recons_1b}]  {\includegraphics[width=0.45\textwidth]{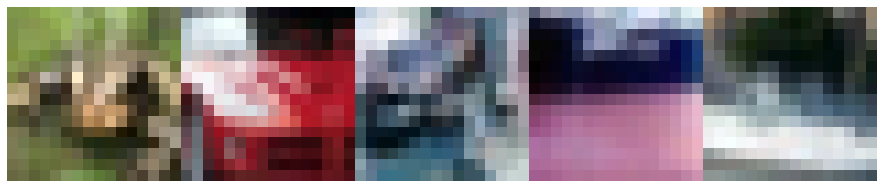}}\hfill
\subfloat[S-LCA Reconstructions with Spike Height of 4. RMSE = 0.0779.  \label{fig:cifar10_recons_1c}]{\includegraphics[width= 0.45\textwidth]{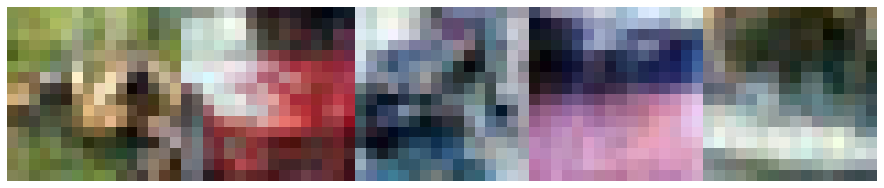}}\hfill
\subfloat[S-LCA Reconstructions with Spike Height of 4 and low-pass filter (100ms). RMSE = 0.0167. \label{fig:cifar10_recons_1d}]{\includegraphics[width=0.45\textwidth]{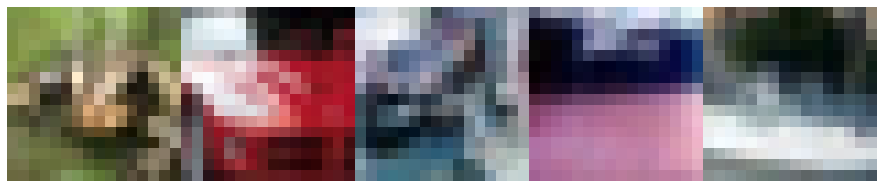}}
\caption{The original 16x16 RGB CIFAR10 images (top row), the reconstructions by LCA (1st middle row), the reconstructions by \hbox{S-LCA} model using accumulator neurons with $s$ = 4 (2nd middle row) and the reconstructions by \hbox{S-LCA} model using accumulator neurons with $s$ of 4 and 100ms low-pass filter(bottom row).} \label{fig:cifar10_recons}
\end{figure}

We measured the quality of the center cropped 16x16 CIFAR10 reconstructions by using the root mean square error (RMSE) metric with image data between 0 and 1. 
Figure~\ref{fig:cifar10_recons} shows examples of original and reconstructed images based on the trained dictionaries from the non-spiking \hbox{LCA} using rate-coded neurons to the \hbox{S-LCA} model using accumulator neurons with spike height of 4. Over 500 images, the two models had similar sparsity values of 1.82\% and 2.17\%, respectively, as measured by average number non-zero coefficient per image.
After normalizing both the original and reconstructed images between 0 and 1,  sparse reconstructions generated by non-spiking LCA had an RMSE of 0.0195 whereas sparse reconstructions generated by S-LCA using accumulator neurons had an RMSE of 0.0779,  which was reduced to an RMSE of 0.0167 after applying a 100ms low-pass filter to the sparse latent representations to temporally average over individual spike events. 
Our results indicate that S-LCA using accumulator neurons with spike height of 4 and a 100ms first-order low pass filter yields lower values of RMSE compared to the non-spiking \hbox{LCA} using rate-coded neurons. The origin of this improvement could be related to the slightly lower sparsity of the S-LCA model.

\subsection{Silicon Retina DVS Camera Events}

\subsubsection{Unsupervised Spatiotemporal Feature Learning}
\hfill \\

To evaluate the performance of S-LCA using accumulator neurons on silicon retina DVS camera events, we trained spatiotemporal features in an unsupervised manner on the Poker-DVS dataset. 
The Poker-DVS dataset \cite{serrano2015poker} is made by fast browsing of a poker deck in front of a DVS event camera. The dataset consists of 4 card types: spades, hearts, diamonds, and clubs. For this work, we created each video frame by accumulating over 1 ms events per card, as illustrated in figure \ref{fig:pokerdvs_input}.


We created video sequences of frames using a 5 frame window within the same Poker-DVS class. By requiring both LCA and S-LCA to optimize features for sparse reconstruction of all 5 frames, we anticipated the dictionary would learn spatiotemporal features corresponding to the the target classes even though learning was unsupervised.
We measure the performance of both LCA and S-LCA by the root mean squared error (RMSE) of the sparse reconstruction on holdout test images. Additionally, we train a perceptron classifier on the sparse latent representations inferred by the two LCA models on the training set. We then capture the corresponding latent representations on the 5-frame validation video sequences and evaluate classification accuracy.

For non-spiking LCA, we trained the model for 30 epochs until the dictionary was roughly stable, showing each image for 100ms, with the input set to zero for 100ms before each image. We tuned the LCA threshold across the range 0.4 to 1. For S-LCA implemented with accumulator neurons, we used the best threshold value from the non-spiking LCA model, and trained the S-LCA model for 10 epochs at various spike heights. As the spike height increases, the maximum number of spikes per time-step decreases from 70 at $s=1$ to 1 at $s=20$.  At the maximum spike height of 20, the accumulator neurons behave close to LIF neurons, typically firing less than 1 time per time-step. We also implemented a moving average to smooth the sparse latent representations inferred by S-LCA. We use the moving average for calculating updates to the dictionary, computing sparse reconstructions for comparison with non-spiking LCA, and for training the linear perceptron classifier. The dynamics of the S-LCA algorithm other than dictionary updates employed no low-pass filtering and used only the spiking output of accumulator neurons. 

\begin{figure}[htp]
\centering
\subfloat[Initial state.\label{fig:poker_dvs_1a}]{\includegraphics[width=0.15\textwidth]{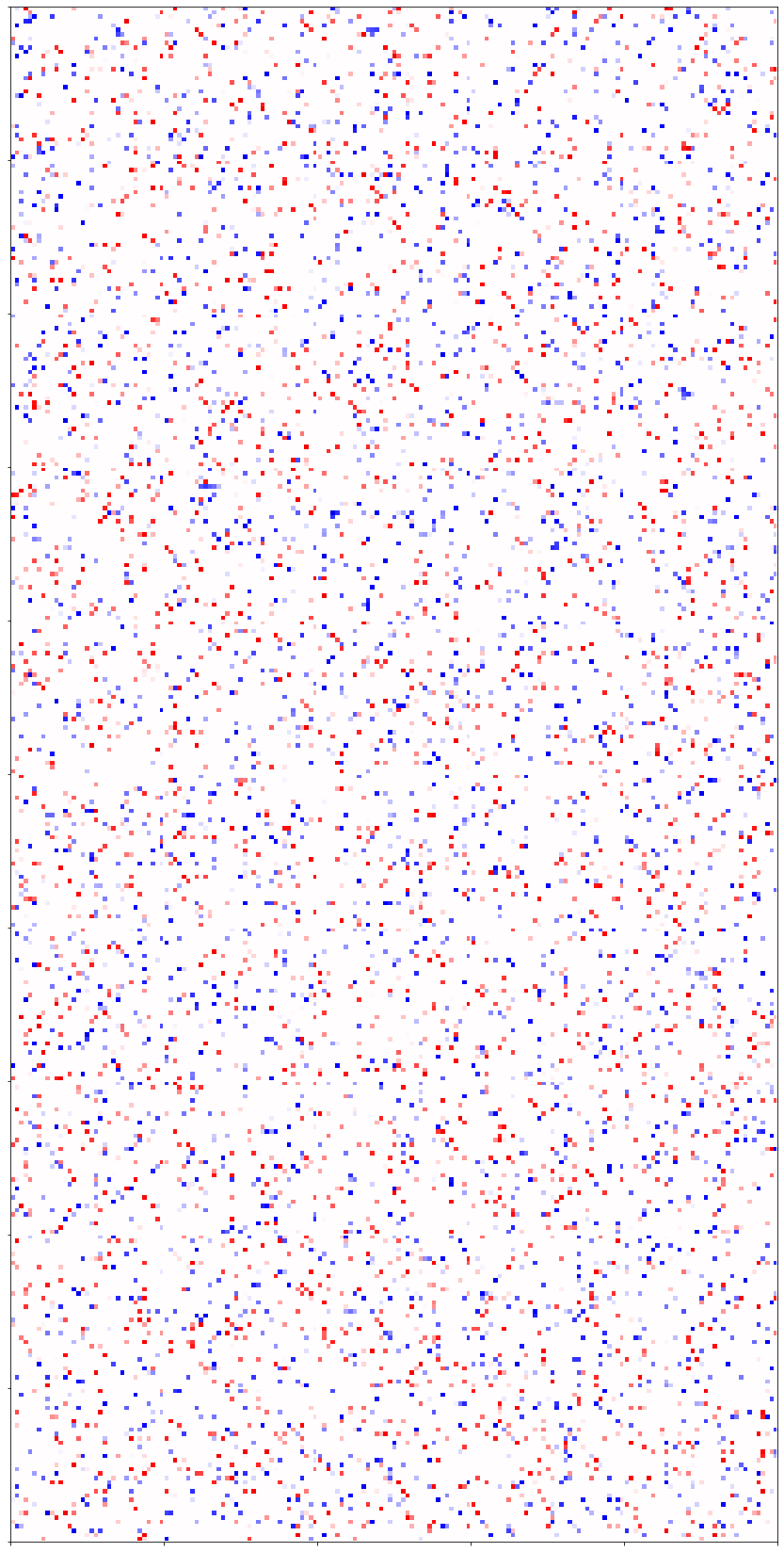}}\hfill
\subfloat[Middle state.\label{fig:poker_dvs_1b}] {\includegraphics[width=0.15\textwidth]{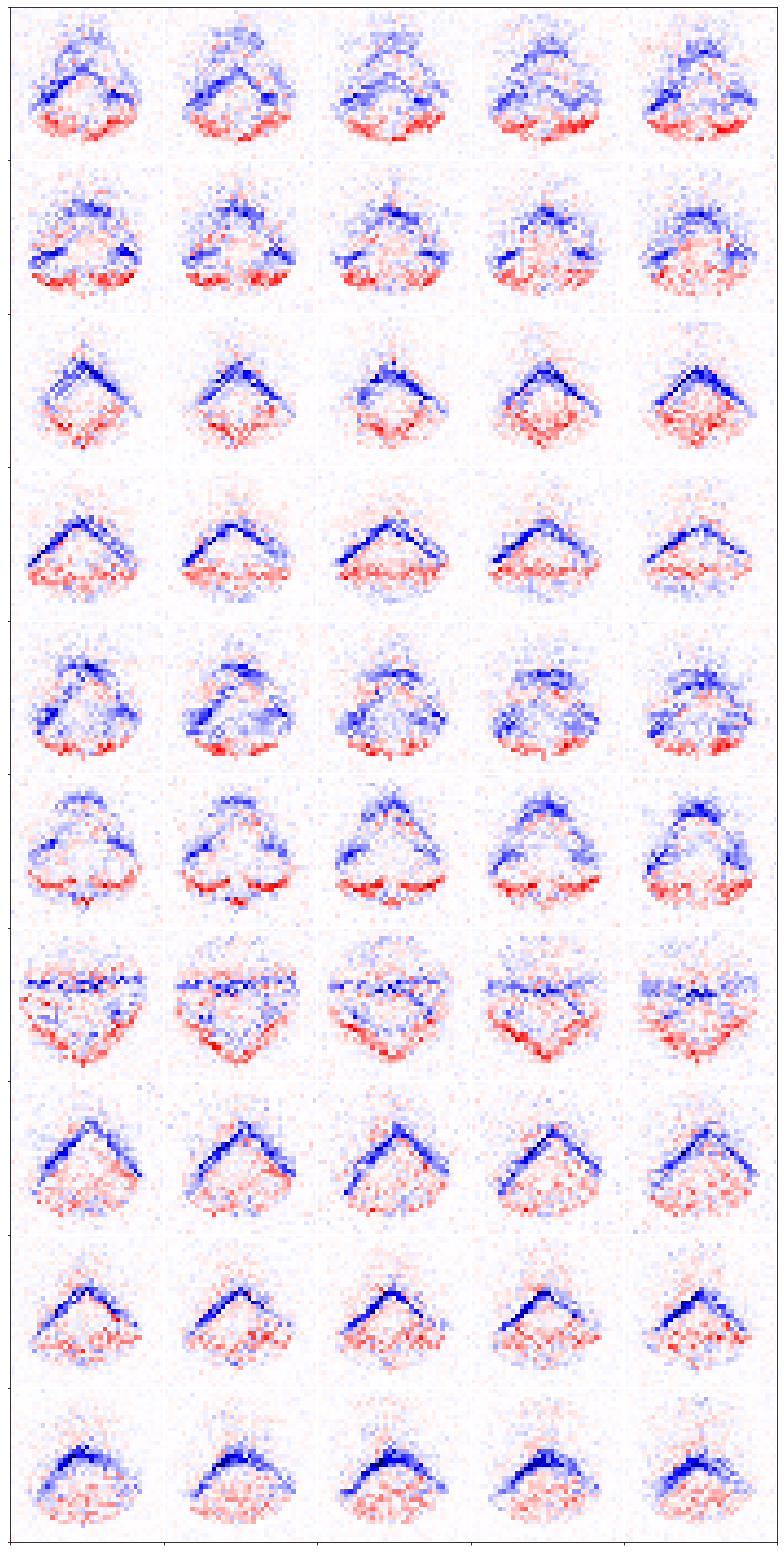}}\hfill
\subfloat[Final state.\label{fig:poker_dvs_1c}]{\includegraphics[width=0.15\textwidth]{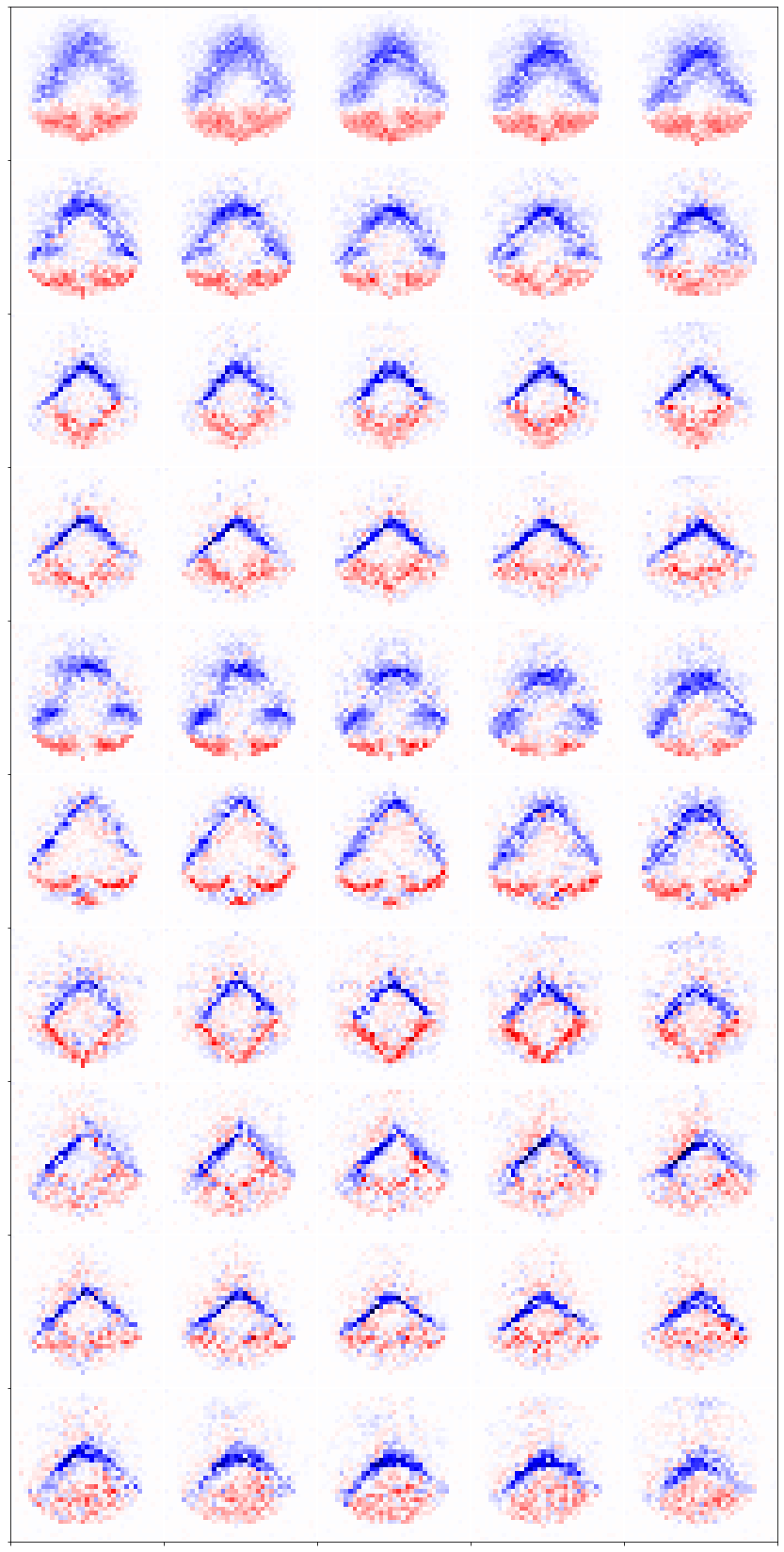}}
\caption{Top 10 most active dictionary elements using the LCA model. Each row is a time-flattened dictionary element. The dictionary elements are randomly initialized in its initial state whereas the final state is how the dictionary elements converge at the last epoch} \label{fig:poker_dvs_original_lca}
\end{figure}

\begin{figure}[htp]
\centering
\subfloat[Initial state.\label{fig:poker_dvs_spiking_1a}]{\includegraphics[width=0.15\textwidth]{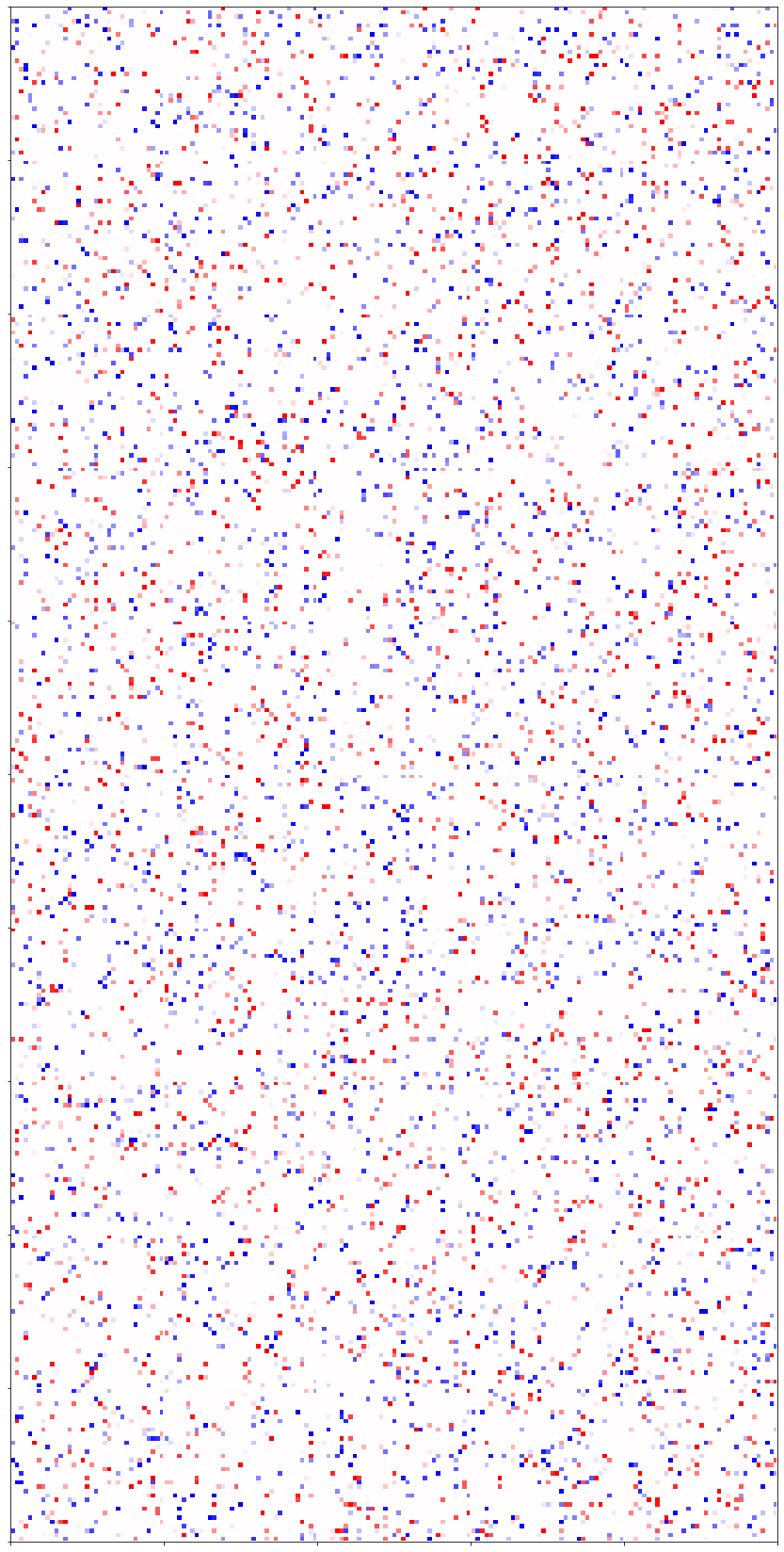}}\hfill
\subfloat[Middle state.\label{fig:poker_dvs_spiking_1b}] {\includegraphics[width=0.15\textwidth]{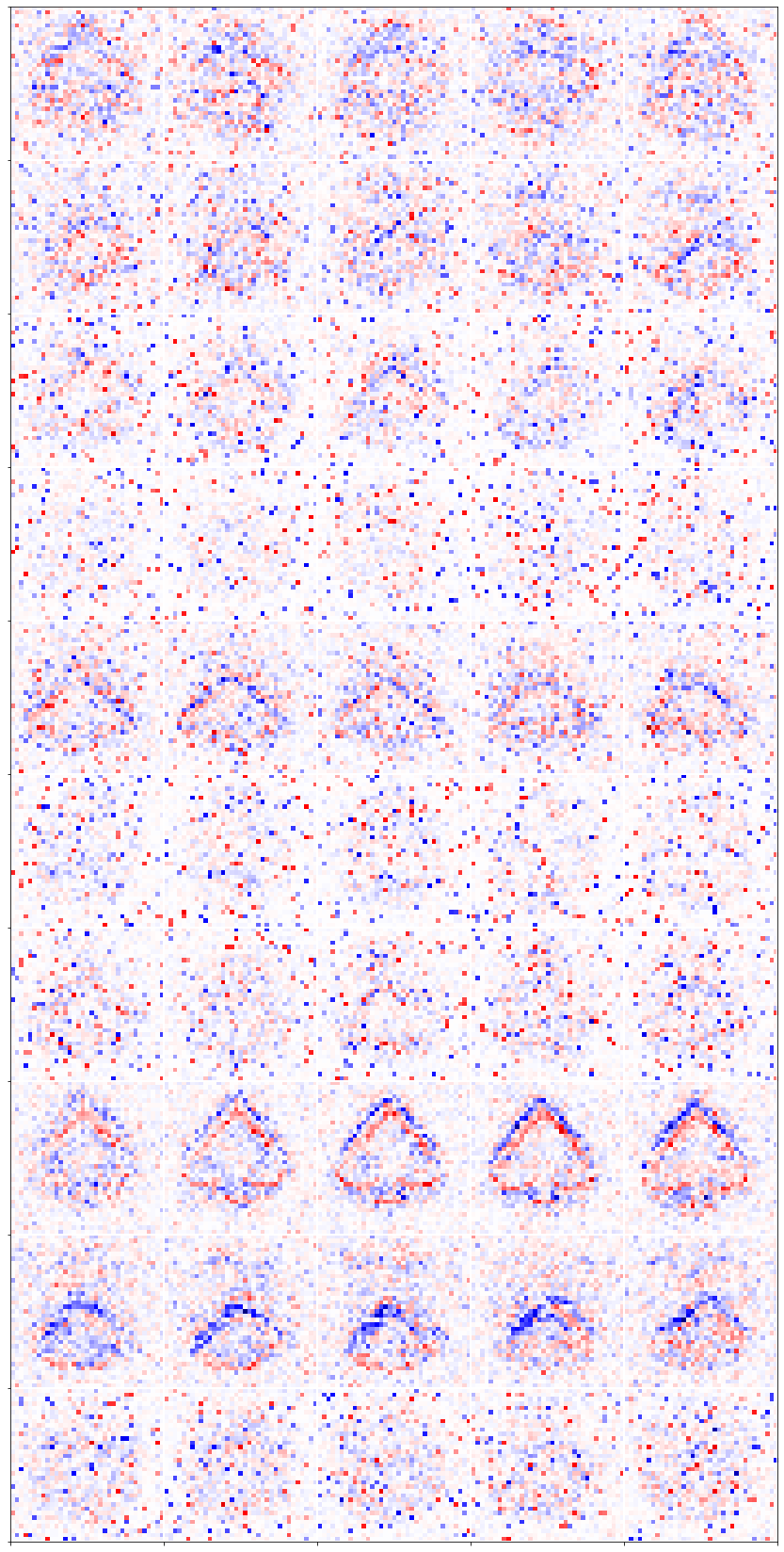}}\hfill
\subfloat[Final state.\label{fig:poker_dvs_spiking_1c}]{\includegraphics[width=0.15\textwidth]{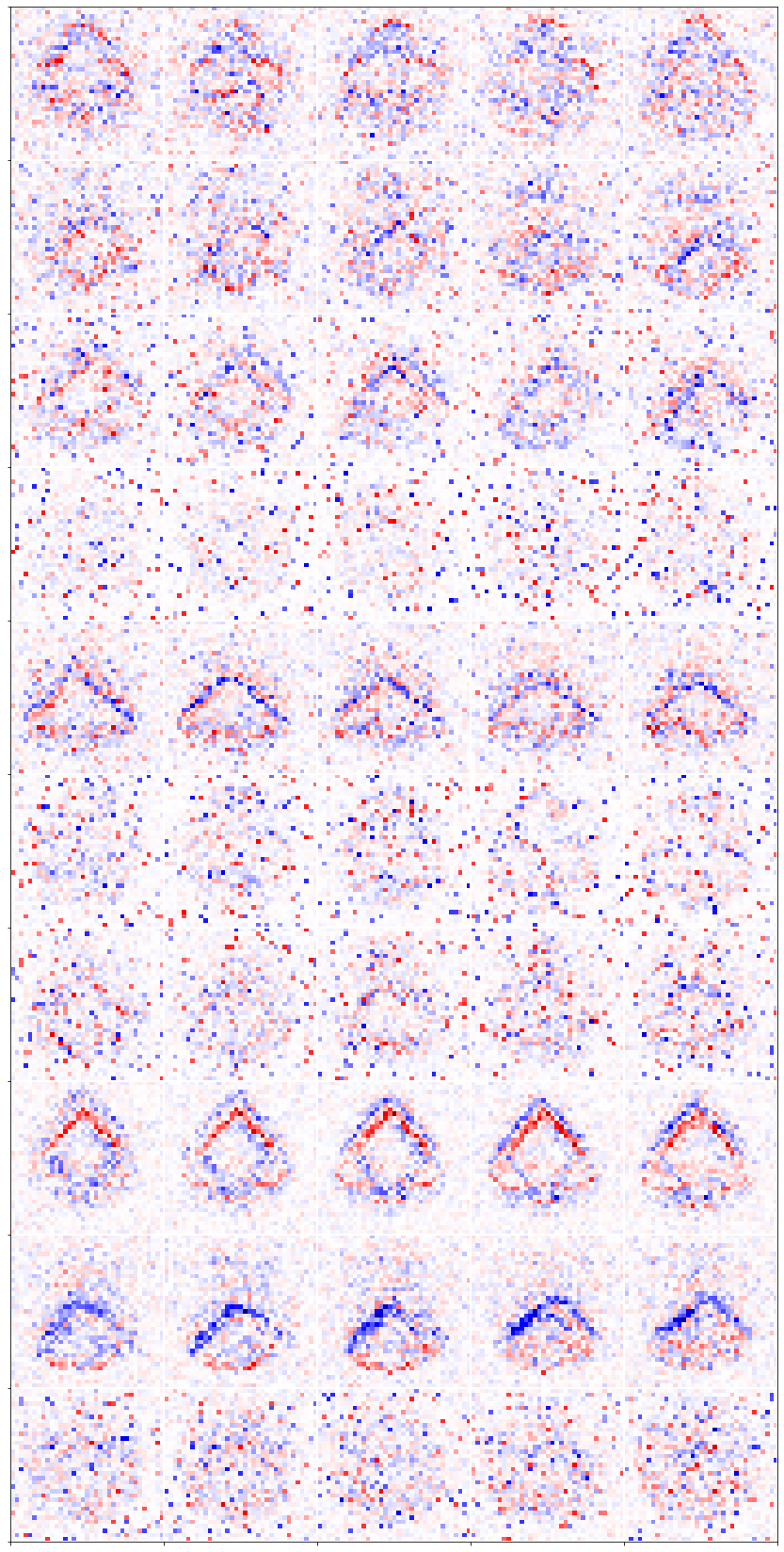}}
\caption{Top 10 most active dictionary elements using S-LCA of $s=10$. Each row is a time-flattened dictionary element. The dictionary elements are randomly initialized in its initial state whereas the final state is how the dictionary elements converge at the last epoch} \label{fig:poker_dvs_spiking}
\end{figure}

The evolution of the dictionary learning over 30 epochs using non-spiking LCA is illustrated in Figure \ref{fig:poker_dvs_original_lca}. 
The evolution of the dictionary learning using accumulator neurons with spike height set to 10 over 30 epochs is illustrated in Figure \ref{fig:poker_dvs_spiking}.

\begin{figure}[htp]
\centering
\subfloat[Initial state.\label{fig:poker_dvs_spiking_avg_1a}]{\includegraphics[width=0.15\textwidth]{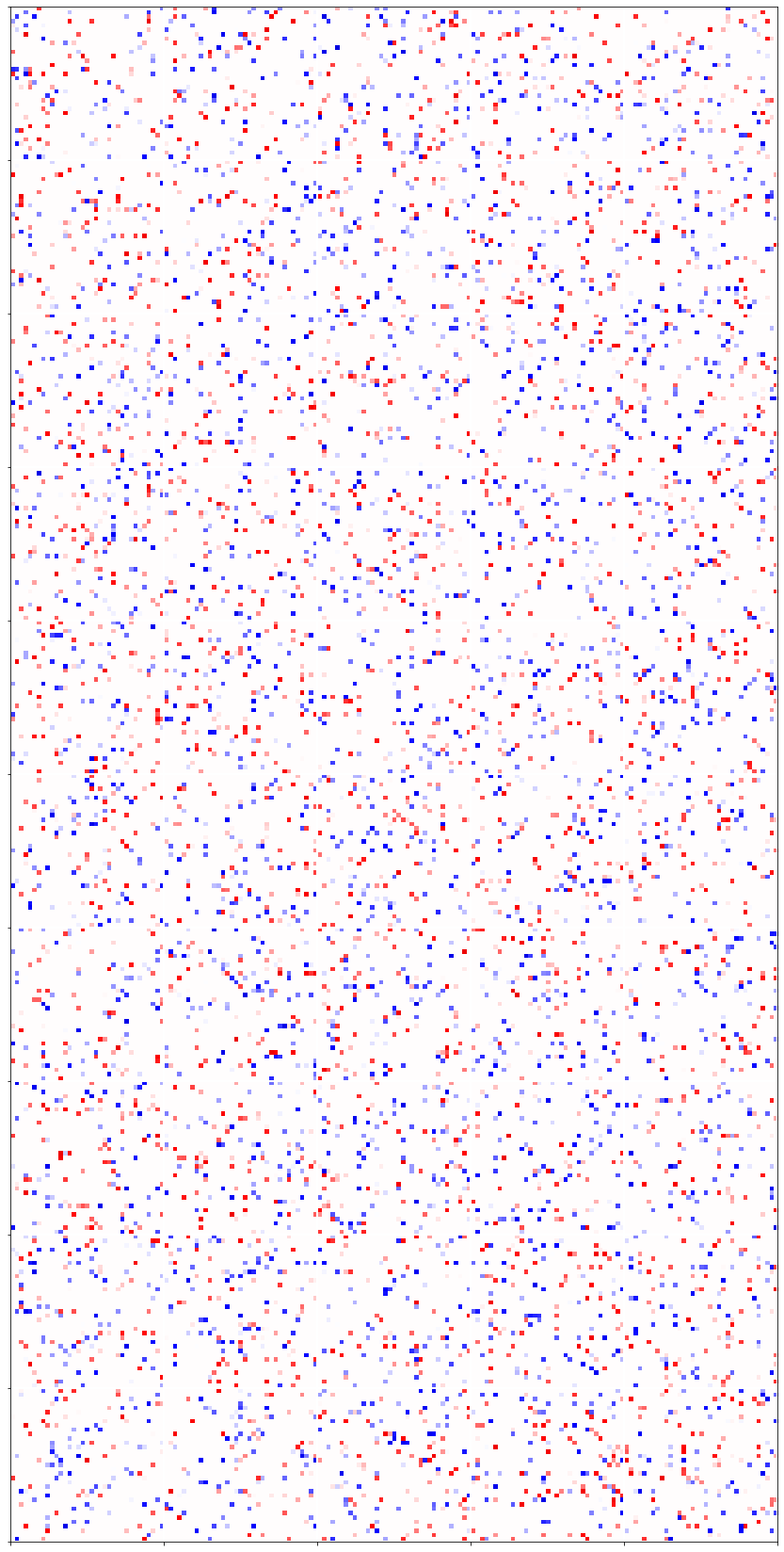}}\hfill
\subfloat[Middle state.\label{fig:poker_dvs_spiking_avg_1b}] {\includegraphics[width=0.15\textwidth]{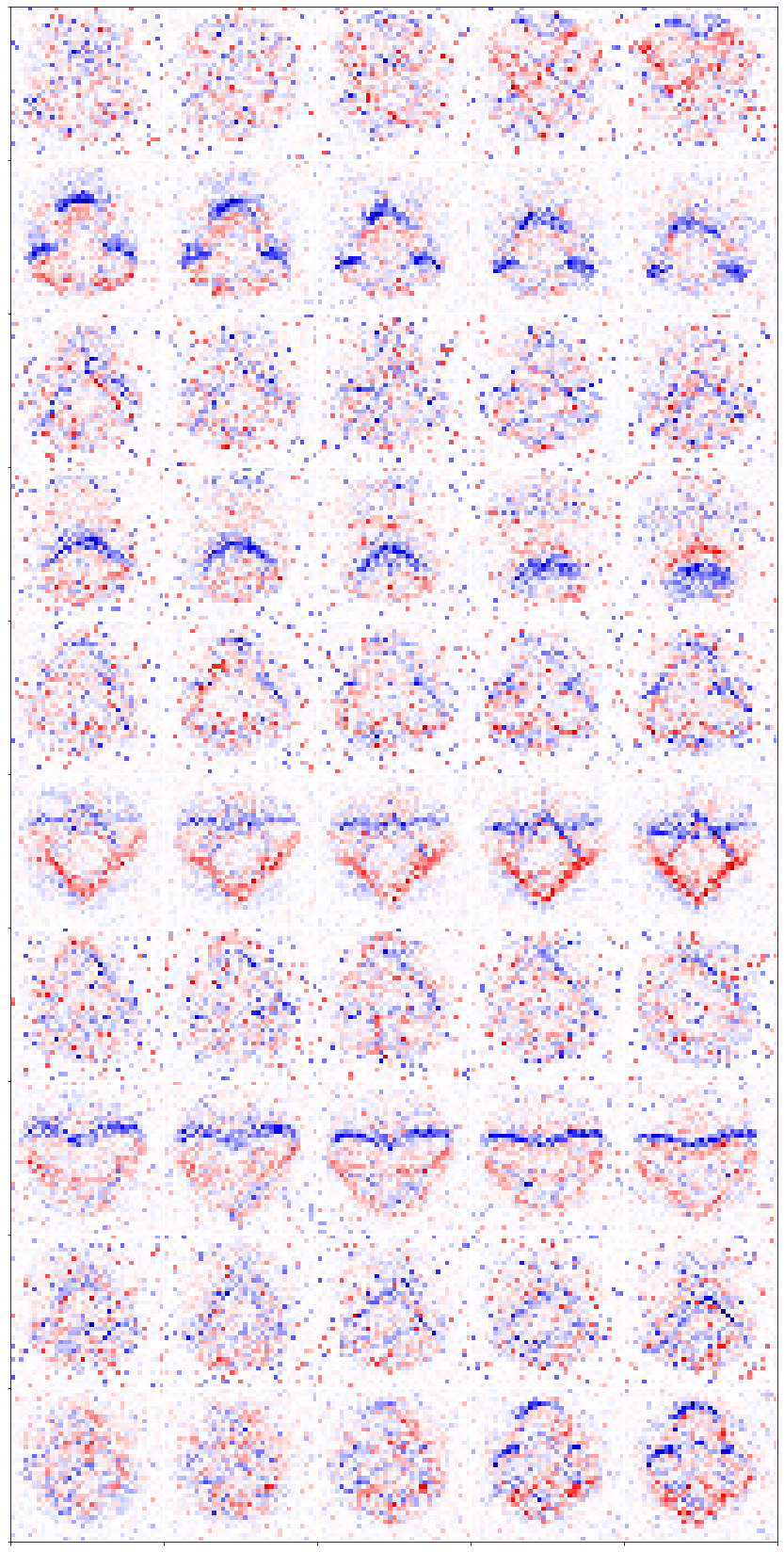}}\hfill
\subfloat[Final state.\label{fig:poker_dvs_spiking_avg_1c}]{\includegraphics[width=0.15\textwidth]{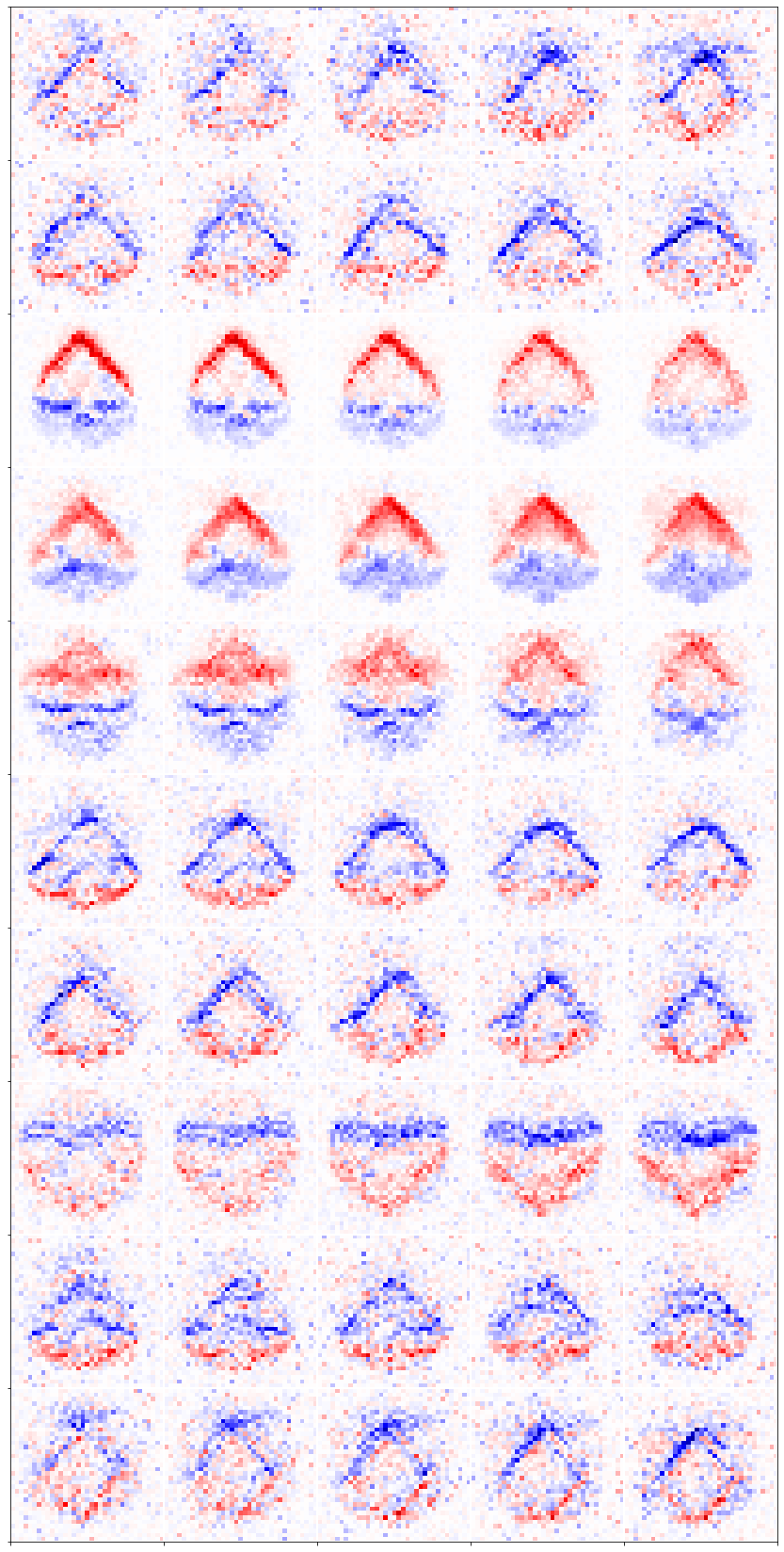}}
\caption{Top 10 most active dictionary elements using S-LCA of $s=10$ with 40 ms averaging applied as a low-pass filter for reconstructions. Each row is a time-flattened dictionary element. The dictionary elements are randomly initialized in its initial state whereas the final state is how the dictionary elements converge at the last epoch} \label{fig:poker_dvs_spiking_avg}
\end{figure}

\subsubsection{Spatiotemporal Sparse Reconstructions}
\hfill \\

\begin{figure}[htp]
\centering
\subfloat[Inputs.\label{fig:dvs_1a}]{\includegraphics[width=0.5\textwidth]{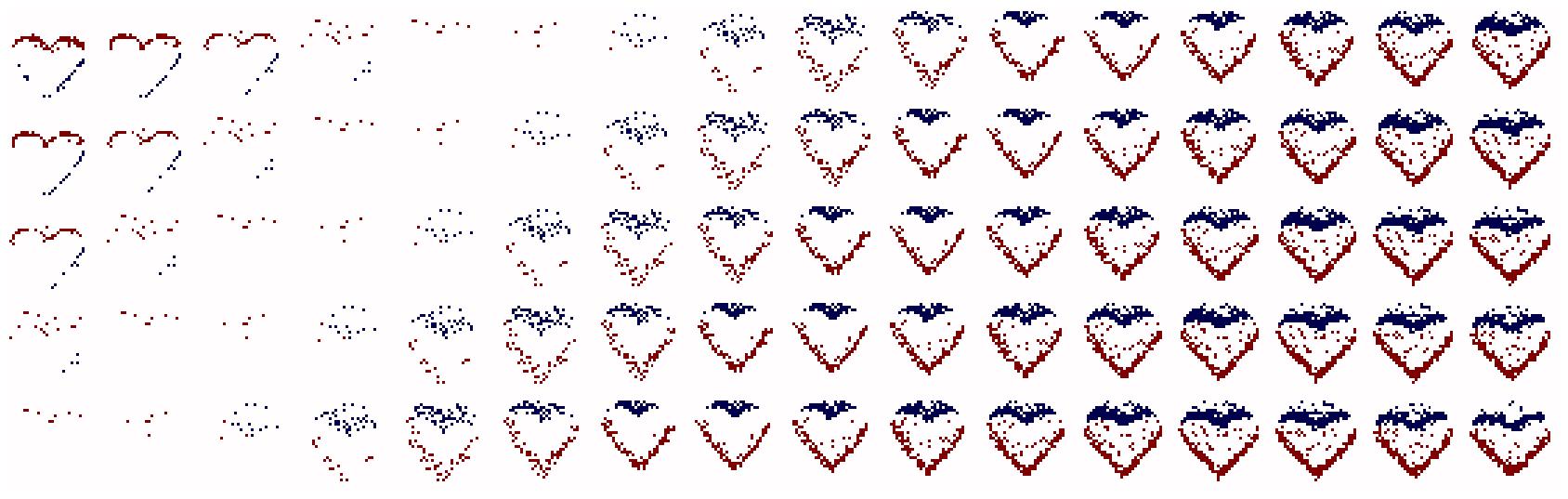}}\hfill
\subfloat[LCA Reconstructions. RMSE = 0.3530 \label{fig:dvs_1b}] {\includegraphics[width=0.5\textwidth]{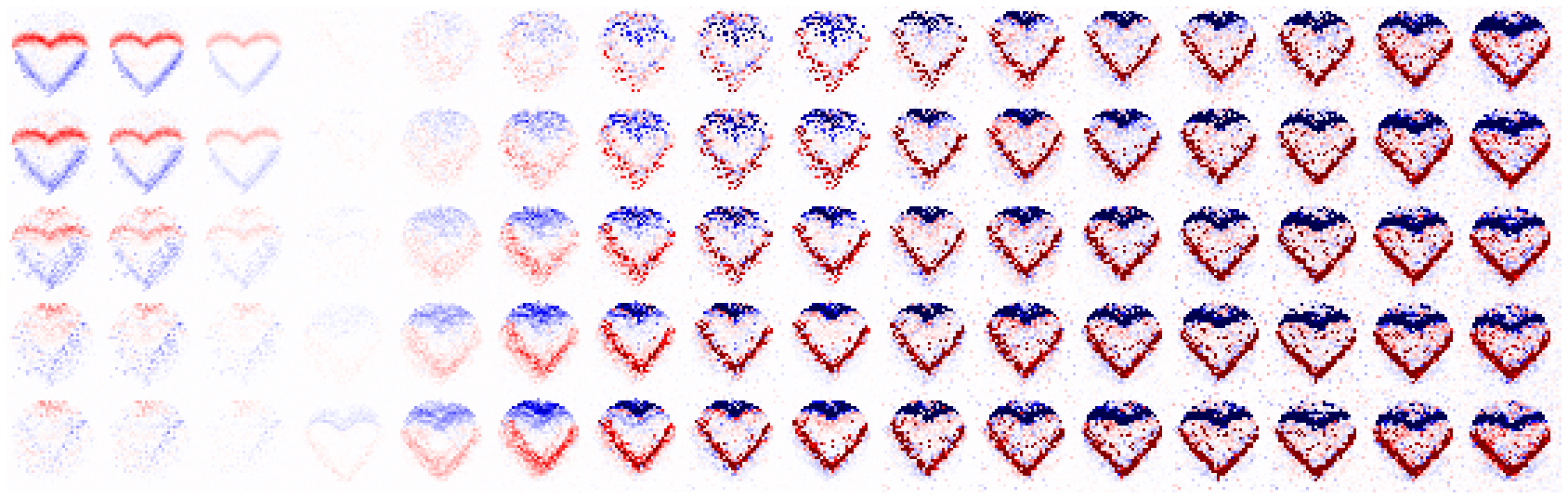}}\hfill
\subfloat[S-LCA Reconstructions $s=10$. RMSE = 0.6405 \label{fig:dvs_1c}]{\includegraphics[width=0.5\textwidth]{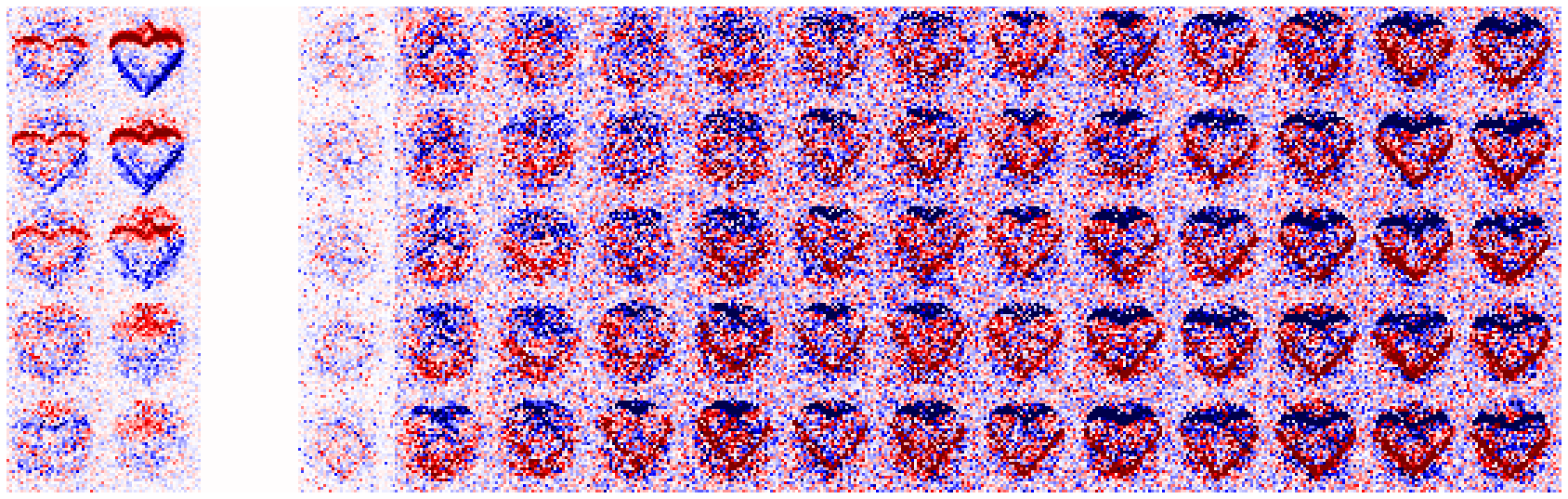}}\hfill
\subfloat[S-LCA Reconstructions $s=10$ with 40ms Moving Average. RMSE = 0.3592 \label{fig:dvs_1d}]{\includegraphics[width=0.5\textwidth]{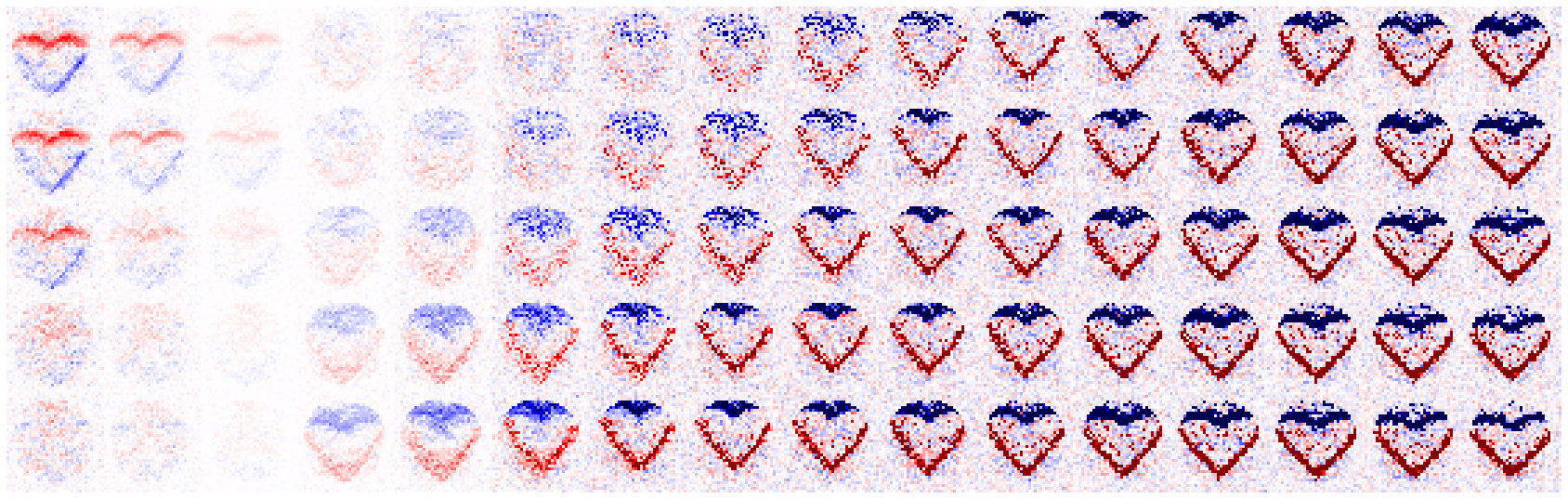}}
\caption{The original Poker-DVS frames (top row), the reconstructions by LCA (second row), the reconstructions by S-LCA using accumulator neurons with $s=10$ (third row), and using $s=10$ with a 40 ms moving average applied (bottom row). The x-axis on each of the rows discretizes between each input while the y-axis shows every individual frame that forms part of the input window}

\label{fig:poker_recons}
\end{figure}

The LCA model was able to learn a dictionary effectively on the DVS frames, with the most used elements shown in Figure \ref{fig:poker_dvs_original_lca}.  The most used elements converged towards classes over time. Figure \ref{fig:poker_recons} shows the reconstructions for various spike heights and RMSE is detailed in Table \ref{tab:dvs_mse_table} with image data between -1 and 1. Similar to CIFAR10, as the spike height increased, the reconstruction accuracy decreased but returned to the performance of non-spiking LCA with the inclusion of a low-pass filter.

\begin{table}[]
    \centering
    \begin{tabular}{c|c|c|c}
    $s$ & No Averaging & 40ms Moving Avg. & Sparsity of 40ms Avg.\\
    1 & 0.3788 & 0.4012 &  90.23\% \\
    5 & 0.4810 & 0.3629 & 86.62\% \\
    10 & 0.6405 & 0.3592 & 87.40\% \\
    20 & 0.9958 & 0.3558 & 85.51\% \\
\end{tabular}
    \caption{Validation RMSE for various spike heights with image data between -1 and 1. All runs with a v1 threshold of 0.6 and a dictionary size of $1/2$.}
    \label{tab:dvs_mse_table}
\end{table}


\subsubsection{Classification}
\begin{figure}[htp]
\includegraphics[width=0.5\textwidth]{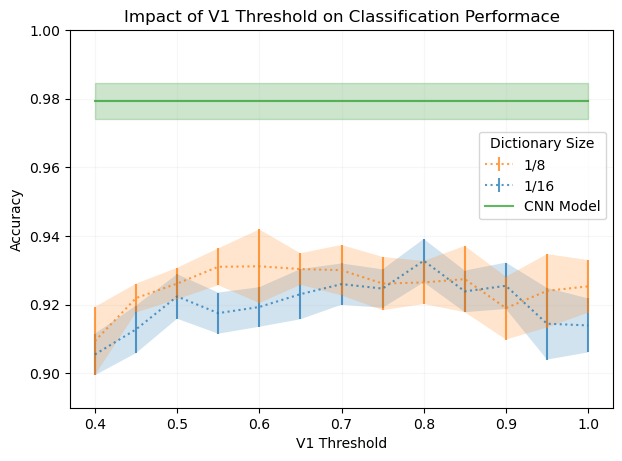}
\caption{Average validation accuracy for LCA with various threshold. Averages are across 10 runs with 95\% confidence intervals shown.}
\label{fig:classification_V1}
\end{figure}

\hfill \\
We focused on the Poker-DVS dataset for classification due to its smaller size and complexity relative to CIFAR10. To evaluate the impact of the LCA threshold and dictionary size on classification performance, we passed the sparse latent representations inferred by non-spiking LCA to a linear perceptron. LCA thresholds ranged from 0.4 to 1 for dictionary sizes of $1/8$ and $1/16$ times under-complete. As a baseline, we compared classification performance obtained using a single hidden layer of sparse latent representations passed to a linear perceptron to the classification performance achieved by a 3-Layer CNN using identical training and validation splits. Each layer consists of a 3D convolution with kernel size 3, 3D batch normalization, ReLu activation, and a 3D max pooling with kernel size 2. The results are shown in Figure \ref{fig:classification_V1}. We observed that the accuracy fluctuated substantially $(\pm ~2\%)$ with different initial weights and so we averaged results across 10 runs. Overall LCA thresholds between 0.6 to 0.8 had higher accuracy on average. Increasing the dictionary size from $1/16$ to $1/8$ also improved accuracy slightly, but further increasing it to $1/2$ with an LCA threshold of 0.6 did not impact accuracy ($92.2\% \pm 0.8$). 

\begin{figure}[htp]
\includegraphics[width=0.5\textwidth]{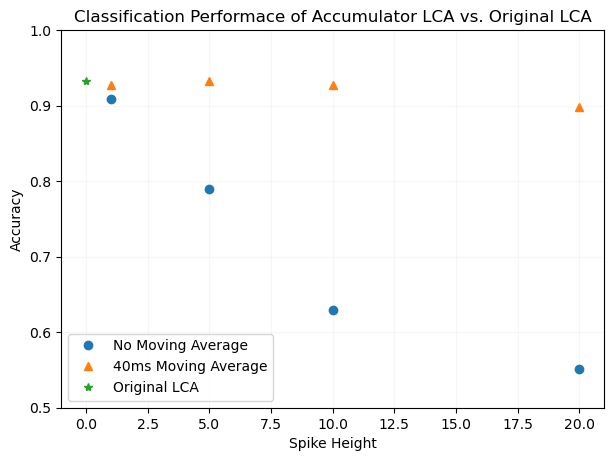}
\caption{Single run validation accuracy for LCA and S-LCA based on accumulator neurons with and without a moving average that smoothed over individual spike events. All models used a dictionary size of $1/2$ under-complete and an LCA threshold of 0.6.}
\label{fig:acc_classification}
\end{figure}

Classification results for S-LCA based on accumulator neurons are depicted in Figure \ref{fig:acc_classification}. Overall accuracy dropped substantially as spike height increased, but adding a moving average to the sparse latent representations inferred by S-LCA dramatically mitigated this impact. At a spike height of 20, where accumulator neurons behave much like a "single spike" LIF neuron, the accuracy is still comparable to non-spiking LCA.

\hfill \\

\hfill \\

\section{Conclusion}

In this work, we demonstrate efficient interpolation from non-spiking to spiking \hbox{LCA} models using accumulator neurons and find essentially no degradation in both sparse reconstruction quality and classification performance.
First, we demonstrate how unsupervised dictionary learning using a locally competitive algorithm \hbox{LCA} can shift from rate-coded neurons to accumulator neurons on both CIFAR10 static images and Poker-DVS video frames with essentially no hand tuning of parameters during the interpolation. 
Second, we find that sparse reconstruction remains at the same quality level when interpolating from non-spiking to spiking S-LCA models employing accumulator neurons whether employing static or spatiotemporal dictionaries. 
Furthermore, we observe no degradation in classification performance when migrating from a non-spiking to a spiking LCA model. Moreover, the classification performance of a single unsupervised hidden layer based on a sparse latent representation passed to a linear perceptron is competitive with a moderately deep, fully-supervised CNN classifier.
Importantly, a classifier based on \hbox{LCA} followed by a linear perceptron requires no backprop and allows for unsupervised dictionary learning, reconstruction, and classification to be performed in a manner compatible with neuromorphic architectures, such as the Intel Loihi research chip.

Unsupervised dictionary learning via sparse coding accounts for many aspects of cortical development. Our results suggest accumulator neurons provide a powerful enabling strategy for endowing future neuromorphic processors with the ability to achieve online unsupervised learning of spatial temporal dictionaries and object recognition and sparse reconstructions of steaming DVS events. 
Future investigations might explore how to process video frames captured from shorter windows that consist of fewer DVS events, in order to utilize the spatiotemporal information available on very short time scales. Future investigations might additionally explore how over-complete dictionaries might permit the learning of more discriminative spatiotemporal features.  The effect on classification performance of using higher spike heights and replacing accumulator neurons with simpler LIF neurons, which can generate only single-bit spikes per time step, is also a promising direction future research.

In summary,  our results suggest that accumulator neurons can be an important enabling component for online unsupervised learning with event based DVS cameras that provide streaming video to ultra-lightweight and low-power neuromorphic processors.

\section{Acknowledgments}
We gratefully acknowledge support from the Advanced Scientific Computing Research (ASCR) program office in the Department of Energy’s (DOE) Office of Science, award no.77902.

\bibliographystyle{ACM-Reference-Format}
\bibliography{citations.bib}

\end{document}